\pdfoutput=1
\documentclass[11pt]{article}

\usepackage[final]{acl}
\usepackage{times}
\usepackage{latexsym}

\usepackage[T1]{fontenc}
\usepackage[utf8]{inputenc}

\usepackage{microtype}

\usepackage{inconsolata}

\usepackage{graphicx}

\usepackage{enumitem}
\usepackage{multirow}  % For multi-row entries
\usepackage{siunitx}   % Align numbers in columns properly
\usepackage{xcolor}
\usepackage{listings}
\usepackage{float}
\usepackage{array}    % For custom column alignment
\usepackage{caption}  % To adjust caption formatting
\setlist[itemize]{topsep=0pt, partopsep=0pt, parsep=0pt, itemsep=0pt}
\usepackage{booktabs, cellspace}

\usepackage{cuted}
\usepackage{hyperref}

\usepackage{titlesec}
\usepackage[symbol]{footmisc}

\newcommand{\adjustimg}{% Horizontal adjustment of image
  \hspace*{\dimexpr\evensidemargin-\oddsidemargin}%
}
\newcommand{\centerimg}[2][width=\textwidth]{% Center an image
  \makebox[\textwidth]{\adjustimg\includegraphics[#1]{#2}}%
}

\let\oldquote\quote
\let\endoldquote\endquote
\renewenvironment{quote}
  {\vspace{-0.3ex}\oldquote\vspace{-0.3ex}}
  {\endoldquote\vspace{-0.3ex}}

\newcommand{\fd}{\textsc{FollowupQG}}

\usepackage{tabularx}
\title{Bridging Information Gaps with Comprehensive Answers:\\Improving the Diversity and Informativeness of Follow-Up Questions}

\author{
  Zhe Liu$^1$\textsuperscript{*}~~~Taekyu Kang$^1$\textsuperscript{*}~~~Haoyu Wang$^1$~~~Seyed Hossein Alavi$^{1,2}$~~~Vered Shwartz$^{1,2}$ \\
  $^1$ University of British Columbia~~~$^2$ Vector Insitute \\
  \texttt{\{zheliu92, salavis, vshwartz\}@cs.ubc.ca} \\
  \texttt{\{davidk15, macdude\}@student.ubc.ca} \\
}

\begin{document}
\maketitle
\renewcommand{\thefootnote}{\fnsymbol{footnote}}
\footnotetext[1]{Denotes Equal Contribution.}
\renewcommand{\thefootnote}{\arabic{footnote}}
\begin{strip}
  % set this to something that looks good.
  \vspace{-58pt}

  % we use the centerimg macro here instead of \includegraphics to make the text on the rest of this fake figure macro correctly render
    \noindent\centerimg[width=1.05\linewidth]{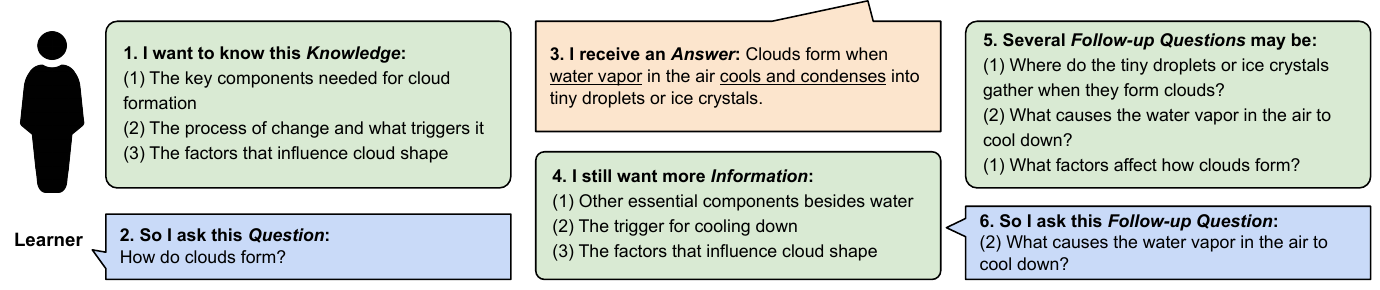}

  % Note: we need to manually write the text "Figure 1:" here as we aren't actually using the "figure" macro.
  \begin{center}
    \fontsize{10pt}{12pt}\selectfont
   Figure 1: Illustration of a learner’s cognitive process in generating follow-up questions. \\ Green: Implicit cognitive stages; Callouts: Explicit (question, answer, follow-up) triplets collected for the dataset.
  \label{fig:teaser}
  \end{center}

\end{strip}

% because we manually made a figure without the macro, we need to increment the figure counter so that the next one is Figure 2
\setcounter{figure}{1}

\begin{abstract}
Generating diverse follow-up questions that uncover missing information remains challenging for conversational agents, particularly when they run on small, locally hosted models. To address this, we develop an \emph{information-gap-driven} knowledge distillation pipeline\footnote{Code available at \url{https://github.com/zheliu92/nlp_followupqg_public/}} in which a teacher LLM generates a comprehensive answer, contrasts it with the initial answer to identify information gaps, and formulates gap-bridging follow-up questions. Using this pipeline, we augment the existing \textsc{FollowupQG} dataset tenfold. We then fine-tune smaller student models on the augmented dataset to distill the teacher’s knowledge. Experiments with selected teacher-student model pairs show that fine-tuned students achieve significantly higher informativeness and diversity than variations trained on the original dataset. These findings indicate that our pipeline, which mirrors the human cognitive process of information seeking, provides an efficient distillation channel from state-of-the-art LLMs to smaller models, enabling resource-constrained conversational systems to generate more diverse and informative follow-up questions.

% Effective conversational systems are expected to dynamically generate contextual follow-up questions to elicit new information. While humans excel at asking diverse and informative questions by intuitively assessing both obtained and missing information, existing models often fall short of human performance on this task. To mitigate this, we propose a pipeline that simulating human's cognitive process to generate diverse and informative questions by targeting unanswered information using a hypothetical LLM-generated ``comprehensive answer''. In this paper, we showed that this pipeline can be applied to augment an existing follow-up questions dataset and achieve 94\% validness. We then demonstrate through experiment that how this pipeline can be used to distill large language model's knowledge into compact language models, through fine-tuning the compact models with the augmented dataset, achieving similarly good performance to the large language model in producing follow-up questions and significantly higer quality and diversity compared to the compact model fine-tuned with the original dataset. The experimental results demonstrate that language models fine-tuned on the augmented datasets produce follow-up questions of significantly higher quality and diversity. \footnote{Code available at \url{https://github.com/zheliu92/nlp_followupqg_public/}}. 

% This promising approach could be effectively adopted to future work to augment information-seeking dialogues for reducing ambiguities and improving the accuracy of LLM answers

\end{abstract}

\section{Introduction}
\label{sec:intro}
Asking questions is a fundamental mechanism for humans to acquire new information, particularly when existing information is incomplete. While large language models (LLMs) excel at passively answering questions from users, their ability to proactively guide conversations by identifying and addressing information gaps remains underdeveloped \cite{liu-etal-2025-superficial}, with smaller models performing even worse. Therefore, the task of question generation (QG) has become a focal point in natural language processing (NLP) for its role in improving information-seeking dialogue systems \cite{chen-etal-2024-learning-retrieve}---including, making information seeking more accurate and efficient \cite{qi-etal-2020-stay}, resolving ambiguities \cite{li2017learning}, and ultimately better understanding users’ needs to provide suitable assistance across various domains \cite{laban-etal-2022-quiz, arslan2024accuracy, li2024mediq}.

While most existing QG tasks focus on generating questions directly answerable from a given context \cite{ zhao-etal-2018-paragraph, pan-etal-2020-semantic, ghanem-etal-2022-question}---a process that diverges from how humans infer and pursue missing information, \citet{meng-etal-2023-followupqg} propose the \fd{} task, which requires models to generate \emph{follow-up questions} that build on, but are not answerable by, the initial question-answer pair. They create the \fd{} dataset and show that existing models often produce repetitive or context-bound questions that fail to target unexplored information \cite{meng-etal-2023-followupqg}. The core challenges of the \fd{} task can be formulated into two dimensions: (1) identifying information gaps, the unanswered aspects of the initial question, and (2) generating diverse questions that target these gaps. 

Building on traditional QG methods \cite{zhao-etal-2018-paragraph, pan-etal-2020-semantic, ghanem-etal-2022-question}, recent work attempts to generate information-seeking follow-up questions using preference optimization \cite{mazzaccara-etal-2024-learning} and knowledge graphs \cite{liu-etal-2025-superficial}, but still lack explicit mechanisms to model gaps or ensure diversity. To address these limitations, we propose an information-gap-driven teacher-student knowledge distillation pipeline. In our approach, a teacher LLM generates a hypothetical ``complete'' response to the initial question, contrasts it with the often incomplete initial answer to identify information gaps, and formulates gap-bridging follow-up questions. By generating multiple follow-up questions, each targeting some unanswered information, this pipeline ensures the diversity and informativeness of the follow-up questions. For example, in Figure \hyperref[fig:teaser]{1}, if the initial answer to ``\textit{how do clouds form?}'' is ``\textit{clouds form when water vapor cools,}'' a comprehensive answer might add ``\textit{… and condenses around dust particles,}'' which explicitly exposes the gap through contrast and leads to an informative follow-up question such as ``\textit{What role do particles play in cloud formation?}''

Our pipeline can be applied across different teacher-student model pairs. In this work, we use GPT-4o (\texttt{2024-02-15-preview}) as the teacher model and BART-large as the student model to verify the pipeline. Specifically, we use GPT-4o to generate the comprehensive answers and follow-up questions. After verifying the quality of the follow-up questions via human evaluation, we then augmented the original \fd{} training set tenfold and fine-tuned BART-large on both the original dataset and our augmented dataset. Leveraging GPT-4o to generate high-quality training data, and then distilling the teacher's knowledge into smaller models, our approach achieves strong performance at a significantly lower cost. The experimental results demonstrate significant improvements of the augmented dataset over the baselines, both in terms of quality (validity, relevance, informativeness, etc.) and diversity. Our contributions are as follows:

\begin{itemize}[leftmargin=10pt]
    \item We propose an \emph{information-gap-driven} teacher-student knowledge distillation pipeline that generates follow-up questions through contrastive analysis of initial answers and generated comprehensive answers.
    \item We augment the \fd{} training set with over 25,000 high-quality synthetic examples.
    \item Experimental results show that small models fine-tuned on our augmented dataset outperform peer small-model baselines and achieve near parity with representative LLM-based \textit{Teacher} and \textit{Chain-of-Thought} models.
\end{itemize}

\section{Related Work}
\label{sec:background}
Question generation (QG) focuses on automatically generating semantically meaningful and well-structured questions based on a given text \cite{ali-etal-2010-automatic}. While traditional QG techniques have made significant strides in domains such as machine comprehension \cite{du-etal-2017-learning, uto-etal-2023-difficulty}, e-commerce \cite{wang2021harvest}, and education \cite{luo-etal-2024-chain}, they primarily generate questions based on known answers. This approach contrasts sharply with human questioning behavior, which actively seeks new information from various perspectives. This limitation has led to the emergence of \fd, a task whose goal is to generate questions that explore previously unanswered or underexplored aspects of a given text.

\fd{} has evolved from simpler methods, such as template-based and retrieval-driven approaches \cite{kumar2017incomplete, soni-roberts-2019-paraphrase, b-etal-2020-automatic}, to more advanced techniques that prioritize informativeness \cite{majumder-etal-2021-ask, mazzaccara-etal-2024-learning}. Knowledge-enhanced approaches, like those in \citet{ge-etal-2023-ask} and \citet{gupta-etal-2022-learning}, leverage entity-relation pairs and knowledge graphs to improve the depth of the generated questions. Further advancing this, \citet{liu-etal-2025-superficial} combined knowledge graphs with LLMs to increase question informativeness. Efforts to model human-like questioning behavior, such as InquisitiveQG \cite{ko-etal-2020-inquisitive}, have relied on crowd-sourced follow-up questions written for news articles rather than those naturally generated by humans, leading to a lack of depth and cognitive diversity.

We follow the setting of the \fd{} \cite{meng-etal-2023-followupqg}, which formalizes information-seeking follow-up question generation. Based on questions and answers from the ELI5 (explain like I'm 5) subreddit, follow-up questions in this dataset build upon---but are not answerable by---the initial question-answer pair, resembling real-world dialogues where follow-ups resolve ambiguities or deepen understanding. 

\begin{figure*}
    \centering
    \includegraphics[width=0.9\textwidth]{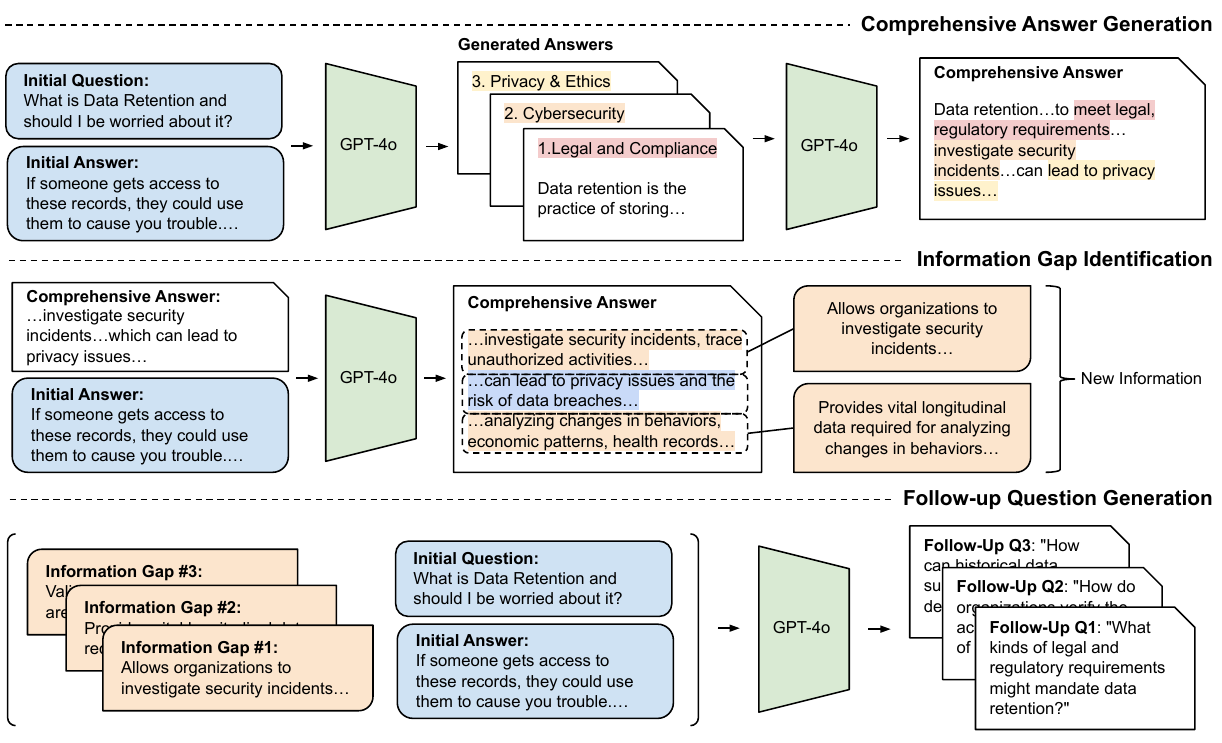}
    \caption{\textbf{Data augmentation pipeline.} For a Q\&A pair, a comprehensive answer is first generated to the question. By comparing it with the initial answer, information gaps are identified. Finally, multiple follow-up questions are generated targeting those gaps.}
    \label{fig:data_aug_pipeline}
\end{figure*}

\newcite{meng-etal-2023-followupqg} found that models often produce questions that are either repetitive or fail to target unexplored information, thus lacking the cognitive diversity and variability seen in human questioning strategies \cite{sultan-etal-2020-importance}. While follow-up QG has made significant progress, existing approaches largely focus on generating questions directly, using various model architectures and knowledge enhancement techniques \cite{ge-etal-2023-ask, liu-etal-2025-superficial}. Our work, however, takes a novel approach inspired by the human cognitive process that models information gaps and uses them to guide the follow-up question generation. 
% through comprehensive answer generation. This step serves as a foundation for follow-up question generation and shifts the focus from comparing model architectures to simulating human cognitive processes in questioning behavior. 
% As a result, we do not compare our method directly with existing models; instead, we evaluate its effectiveness by analyzing the informativeness and diversity of the questions generated by multiple models fine-tuned on variants of the \fd dataset.

\section{Data Augmentation}
\label{sec:data}
Effective \fd{} requires models to infer and target gaps between the provided answer and the broader context of a conversation. Following the task definition by \citet{meng-etal-2023-followupqg}: ``to generate follow-up questions that seek new information given the initial question and answer'', we denote the ``initial question'' as IQ, ``initial answer'' as IA, and the ``follow-up question'' as FQ. We identified critical limitations in the training dataset, including quality issues, which we addressed through dataset cleaning (\S\ref{sec:data:cleaning}). To overcome the small scale (2,790 instances) and low diversity of the dataset, we present a novel data augmentation pipeline (\S\ref{sec:data:augmentation}). Finally, we demonstrate that the augmented dataset retains high quality (\S\ref{sec:data:quality}).

\subsection{Data Cleaning}
\label{sec:data:cleaning}

The \fd{} dataset is limited by its small scale, comprising 3,790 samples: 2,790 for training, 500 for validation, and 500 for testing. Within the 2,790 training instances, there are only 2,651 unique (IQ, IA, FQ) triplets, indicating duplication. Moreover, the dataset consists of 2,648 unique (IQ, IA) pairs, meaning that 99.8\% of the (IQ, IA) pairs have only one reference FQ. Training models on this set could thus lead to poor follow-up question diversity. Our further analysis also uncovered data quality issues, likely stemming from automated data collection (see Appendix~\ref{app:problematic_sample}). To improve the data quality, we did the following:

\begin{itemize}[leftmargin=10pt]
    \item \textbf{Deduplication}: We removed 139 duplicate (IQ, IA, FQ) triplets.
    \item \textbf{Reference quality check}: We manually filtered out 84 instances where the reference FQ diverged entirely from the initial question. 
    \item \textbf{Sensitive content removal}: We excluded 24 instances involving topics like self-harm or crime, which LLMs are likely to refuse to answer.
\end{itemize}

% \paragraph{Deduplication.} We removed 139 duplicate (IQ, IA, FQ) triplet instances. 

% \paragraph{Reference quality check.} We manually filtered out 84 instances where the reference FQ diverged entirely from the initial question. %This step ensures coherence in the augmented dataset.   

% \paragraph{Sensitive content removal.} We excluded 24 instances involving topics like self-harm or crime, which modern LLMs are likely to refuse to answer. %, avoiding pipeline failures during augmentation.

The cleaned dataset (2,543 instances) retained broad topic coverage, containing 2,533 unique (IQ, IA) pairs.

\subsection{Augmentation Pipeline}
\label{sec:data:augmentation}

As discussed in \S\ref{sec:data:cleaning}, the limited scale of the dataset and the lack of follow-up question diversity hinder the coverage of diverse questioning strategies, restricting model generalization. To address this, we design a GPT-4o-based pipeline that augments the original dataset by generating additional follow-up questions. Our pipeline simulates human reasoning through three interconnected stages: comprehensive answer generation, information gap identification, and follow-up question generation.\footnote{Please refer to Appendix~\ref{app:prompts} for the LLM prompts used for the following stages.} %Each stage aims to enhance diversity, improve quality, and align with human strategies for effective information-seeking dialogues.

\paragraph{Comprehensive answer generation.} To identify gaps in the IA, we generate a comprehensive answer (CA) that represents a complete and thorough response to the IQ. As shown in Figure~\ref{fig:data_aug_pipeline}, we prompt GPT-4o iteratively to generate answers to IQ that target different perspectives, and synthesize a unified CA. More specifically, GPT-4o was prompted to generate a combination of new and different answers that do not overlap with the other answers, where each answer focuses on a unique aspect not covered in the other generated answers.
% Prompts such as ``Generate a concise answer focused on a single perspective'' and ``Synthesize prior answers into a comprehensive explanation'' ensure the CA is broad yet cohesive. This process mirrors how humans consider multiple dimensions to intuit missing information. 

\paragraph{Information gap identification.} The next step is to identify key concepts or details discussed in the comprehensive answer (CA) but not covered in the initial answer (IA). This is done by prompting GPT-4o. As shown in Figure~\ref{fig:data_aug_pipeline}, the initial answer covers the topic of privacy issues but does not cover areas of cyber security (i.e. an information gap). To confirm the validity and reasonableness of the comprehensive answers and identified information gaps, we manually evaluated a random sample of comprehensive answers and ensured that they were accurate and reasonable. Examples of comprehensive answers can be seen in Table \ref{tab:example_question_long} and Table \ref{tab:example_question_short}.
%The LLM uses the prompt: ``Identify specific explanations or concepts in the comprehensive answer absent from the initial answer.'' 

\paragraph{Follow-up question generation.} Using the identified information gaps, we prompt GPT-4o to generate follow-up questions that address those gaps while maintaining contextual relevance to the IQ and IA. The generated questions must meet three criteria: be (1) answerable by the CA, (2) unanswerable by the IA, and (3) grounded in terminology and context from the IQ. % This ensures that the follow-up questions are focused, relevant, and factually grounded, emulating human curiosity and reasoning.

% \paragraph{Dataset Reformation and Merging}
After augmentation, each (IQ, IA) pair now includes an average of 10.95 FQs. To preserve the original \fd{} format, we automatically remove artifacts such as bullets or numbering from the generated FQs and merge them with the cleaned human-written examples. The resulting dataset comprises 27,874 samples---about 10$\times$ the original size---and better reflects the open-ended nature of human questioning, providing models with diverse, explicit signals for addressing information gaps.

\subsection{Augmented Data Validation}
\label{sec:data:quality}

To assess the quality of the generated follow-up questions, we conducted a human evaluation study on Cloud Connect, using \citet{meng-etal-2023-followupqg}'s survey. To ensure high-quality annotations, we restricted participation to native English-speaking annotators with a minimum of 1,000 completed annotation tasks and an approval rating exceeding 90\%. A randomly sampled subset of 100 (IQ, IA, FQ) triplets was evaluated based on three key criteria: (1) whether the FQ was a valid question,\footnote{Following \newcite{meng-etal-2023-followupqg}, a valid question must be in a question format and ask for meaningful information, including Wh-questions (what/why/where/etc.), open-ended questions, probing questions, etc.} (2) whether any component of the triplet contained sensitive information, and (3) the degree of relatedness between the FQ and the (IQ, IA) pair. The full survey format, including example annotations, is provided in Appendix~\ref{sec:augmented_data_annotation_guideline}. 
The results show that 94\% of the FQs are labeled as valid, 92\% as not sensitive, and 91\% are related to the original (IQ, IA) pair. Inter-annotator agreement was moderate, with a Cohen's Kappa score of $\kappa = 0.73$ \cite{cohen1960coefficient}.
%, demonstrating strong inter-rater reliability in the evaluation process.

% \label{human_GPT}

\section{Experiment Setup}
\label{sec:exp_setup}
\paragraph{Model Variants.} To assess our proposed pipeline and augmented dataset, we fine-tuned BART-large \cite{lewis-etal-2020-bart} (24 layers, 16 attention heads, hidden size = 1024) on several versions of the \fd{} data \cite{meng-etal-2023-followupqg}, producing three model variants. BART-large is a seq2seq model that conditions on the concatenated IQ and IA to generate an FQ. We chose it as our base model because of its strong performance reported by \citet{meng-etal-2023-followupqg}. As their implementation is not public, we reproduced their training setup (batch = 8, epochs = 10, Adam \cite{kinga2015method}) and found that the original learning rate of 5e-5 caused instability, so we reduced it to 2e-5; all other hyperparameters remain unchanged.\footnote{Full hyperparameter details and reproduction results are provided in Appendix \ref{appendix:baseline_reproduce}.}

\begin{figure}[t]
    \centering
    \includegraphics[width=0.6\columnwidth]{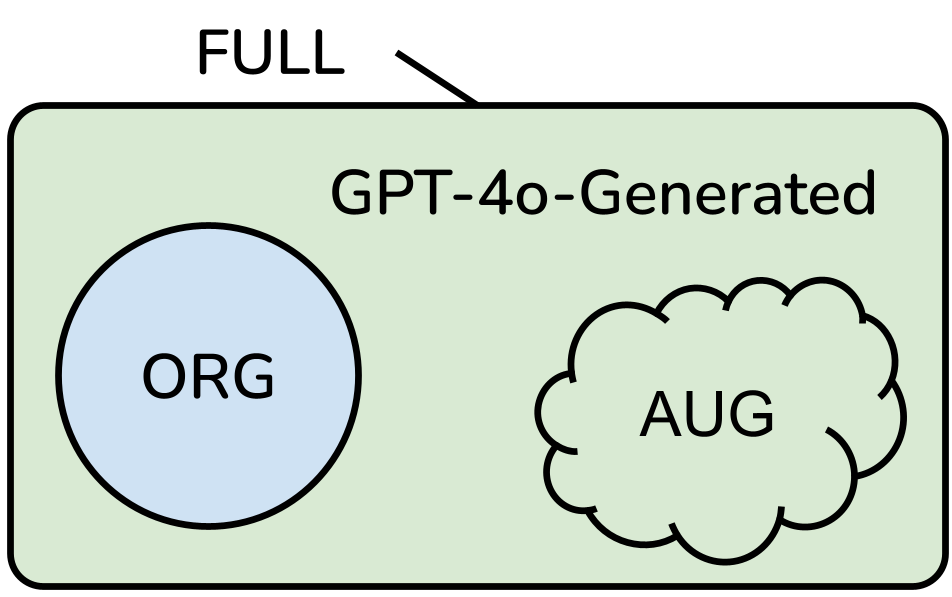}
    \caption{\textbf{The ORG/AUG/FULL Dataset.} ORG is the original dataset of ELI5 from the FollowupQG paper. We augment ORG using GPT-4o. FULL combines ORG with all GPT-4o-generated data (about 10$\times$ larger). AUG is a random sample of only GPT-4o-generated instances, equal in size to ORG and excluding its data.}
    \label{fig:full_aug_org_diagram}
\end{figure}

We report three variants trained on the \fd{} training set as shown in Figure \ref{fig:full_aug_org_diagram}: the \textit{ORG} model is trained on the 2,790 original instances from \citet{meng-etal-2023-followupqg} and serves as our small-model baseline; %\footnote{Two additional large-model baselines are introduced later in the \textbf{LLM Baselines} section}; 
the \textit{AUG} model, trained on a size-matched random sample of 2,790 GPT-generated questions from our augmentation pipeline (Sec.~\ref{sec:data:augmentation}) to isolate the effect of data quality; and the \textit{FULL} model, trained on the entire 27,874-instance augmented dataset. All variants share identical hyperparameters and are evaluated on the original \fd{} validation and test splits.

\paragraph{Decoding.} To generate diverse but contextually relevant follow-up questions, we input the initial question and answer into the model in the following format: \texttt{IQ <SEP> IA}, and generate 10 follow-up questions by applying beam search with a beam width of 20, selecting the top 10 candidates. We added a diversity penalty of 10 to encourage unique outputs across the groups and set the temperature to $t =1.0$ to maintain a balance between diversity and coherence. The maximum length for each generation is set to 1024 tokens. Duplicate generations are removed. 

\paragraph{LLM Baselines.} Beyond the three BART variants, we evaluated two GPT-4o baselines that represent current large-model performance. The \textit{Teacher} setting executes the full information-gap pipeline from \S\ref{sec:data:augmentation} on each test instance, returning GPT-4o’s comprehensive answer, the identified gaps, and the resulting follow-up questions. The \textit{CoT} setting applies the same chain-of-thought prompt but suppresses all intermediate reasoning, outputting only the follow-up questions. Duplicate generations are removed as in the BART variants.

\section{Results}
\label{sec:results}
To thoroughly assess the quality of the generated follow-up questions, we employ both automatic evaluation (\S\ref{sec:results:auto}) and human evaluation (\S\ref{sec:results:human}). As a first step for both evaluations, we automatically identify and remove ungrammatical questions based on syntactic parsing (see Appendix~\ref{appendix:ungrammatical_questions} for a complete description of the filtering process). Table~\ref{tab:filtered} shows the percentage of ungrammatical questions that were filtered out for each model. \textit{AUG} (3.58\%) and \textit{FULL} (6.31\%) produce far fewer ungrammatical FQs compared to \textit{ORG} (33.25\%), demonstrating their ability to generate more well-formed outputs. We focus the rest of our evaluation on the grammatical questions retained after the filtering.

\begin{table}[!t]
    \centering
    \small
    \setlength{\tabcolsep}{6pt} % Adjust column spacing
    \begin{tabular}{lccc}
        \toprule
        \textbf{Model} & \textbf{Total} & \textbf{Ungrammatical} & \textbf{Filtered (\%)} \\
        \midrule
        \textit{ORG}  & 2349 & 781  & 33.25  \\
        \textit{AUG}  & 1895 & 68   & \textbf{3.58}   \\
        \textit{FULL} & 2061 & 130  & 6.31   \\
        \bottomrule
    \end{tabular}
    \caption{Percentage of filtered-out ungrammatical FQs.}
    \label{tab:filtered}
\end{table}

% To address this, we implemented a multi-step validation framework that filters out errors in syntax, semantics, and wording. Using \texttt{spaCy}'s dependency parser, we check for proper WH-question indicators and auxiliary verbs. We also verify the presence of question marks and remove any artifacts or non-informative placeholders. Additionally, we use consecutive n-gram analysis to detect and eliminate excessive duplication between the original question-answer pair and the generated follow-up.

\subsection{Automatic Evaluation}
\label{sec:results:auto}
\begin{table}[!t]
    \centering
    \scriptsize
    \setlength{\tabcolsep}{2pt}  % Tighter column padding
    \begin{tabular}{lccccccc}
        \toprule
        & \multicolumn{3}{c}{\textbf{Diversity}} & \multicolumn{4}{c}{\textbf{Length (in token)}}\\ 
        \cmidrule(lr){2-4} \cmidrule(lr){5-8}
        \textbf{Model} & \shortstack{Distinct-1\\(\%)} & \shortstack{Distinct-2\\(\%)} & \shortstack{Clusters\\per FQ} & Avg. & Shortest & Longest & Std. Dev. \\
        \midrule
        \textit{ORG}  & 66.06 & 91.12 & 0.651 & 14.25 & 3  & 111 & 10.13  \\
        \textit{AUG} & \textbf{77.36} & 94.41 & 0.857 & 13.13 & 4  & \textbf{24}  & \textbf{2.98} \\
        \textit{FULL}  & 77.09 & \textbf{94.85} & \textbf{0.866} & 13.17 & 4  & 73  & 3.77  \\
        \midrule
        \textit{Teacher}  & 77.00 & 95.07 & 0.869 & 11.96 & 3  & 25 & 3.19  \\
        \textit{CoT} & 80.65 & 96.34 & 0.878 & 16.36 & 5  & 30  & 4.02 \\
        \bottomrule
    \end{tabular}
    \caption{Automatic evaluation of follow-up question generation without human reference.}
    \label{tab:auto_eval}
\end{table}

% \textbf{\textit{Vered:I may add it back later but I think it's already implied -- please confirm}}
% \textit{These results highlight the limitations of automatic metrics for evaluating follow-up question generation. Unlike answer-aware QG, where factoid questions closely match reference answers, follow-up questions are open-ended, allowing for multiple valid generations. This lexical diversity lowers traditional n-gram-based scores, suggesting that future evaluations should incorporate human assessments or embedding-based similarity measures to better reflect the quality of generated questions.}

\paragraph{Diversity.} We assess the diversity of each set of FQs at the (IQ, IA) level and average the scores across the dataset. First, we report Distinct-n \cite{li-etal-2016-diversity}, which measures the average distinct n-gram in the FQs associated with each (IQ, IA) pair. Table \ref{tab:auto_eval} shows that \textit{AUG} and \textit{FULL} achieve comparable Distinct-1/2 scores, both exceeding \textit{ORG}. Moreover, the \textit{AUG} Distinct-1/2 scores are comparable to those of the GPT-4o \textit{Teacher} baseline and only slightly lower than those of the advanced \textit{CoT} model. 

We also compute a sentence-level diversity score. We embed the FQs using \texttt{all-mpnet-base-v2} \cite{reimers-gurevych-2019-sentence} and apply agglomerative clustering at a distance threshold of 1.0, normalizing the number of clusters by the number of generated follow-up questions. A score of 1 denotes maximum diversity whereas lower values indicate that questions collapse into the same cluster. 
Again, Table~\ref{tab:auto_eval} confirms the trend that our augmentation substantially improves the diversity. Moreover, for both metrics, \textit{AUG} is statistically indistinguishable from the GPT-4o \textit{Teacher} and \textit{CoT} baselines, showing that our pipeline elevates small models to LLM-level diversity on the \fd{} task.

% \label{sec:length_analysis}
\paragraph{Average question length.} We report the average question length in terms of the number of tokens. We hypothesize that shorter questions are generally more readable. Table~\ref{tab:auto_eval} lists the average length, shortest and longest follow-ups, and standard deviation (SD). The \textit{ORG} model shows the greatest variation in question length (SD = 10.13). Notably, its longest follow-up (111 tokens) far exceeds \textit{FULL} (73) and \textit{AUG} (24). In contrast, \textit{AUG} is the most consistent (SD = 2.98; max = 24), with \textit{FULL} close behind (SD = 3.77)

Qualitatively examining the generated follow-up questions, we find that \textit{AUG} and \textit{FULL} generally produce concise, well-formed queries, while \textit{ORG} sometimes generates very short, vague prompts (e.g., ``So it's cultural?''). Meanwhile, the longer questions from \textit{ORG} and \textit{FULL} often include extraneous conversational filler. Overall, \textit{AUG} maintains structured, concise outputs for follow-up questions, whereas \textit{FULL} and \textit{ORG} exhibit greater variability, occasionally producing overly long or conversational phrasing. More examples are provided in \S\ref{sec:analysis:casestudy} and Appendix~\ref{app:additional_examples}.

\begin{table}[!t]
    \centering
    \scriptsize
    \setlength{\tabcolsep}{3pt}  % Tighter column padding
    % \begin{tabular}{@{}lccccccccccccccc@{}}
    % \scalebox{0.9}{
    \begin{tabular}{lcccccccc}
        \toprule
        % & \multicolumn{3}{c}{\textbf{Diversity}} & \multicolumn{4}{c}{\textbf{Length}} & \multicolumn{8}{c}{\textbf{Reference-Based}}\\ 
        % \cmidrule(lr){2-4} \cmidrule(lr){5-8} \cmidrule(lr){9-16}
        \textbf{Model} & \textbf{BERT} & \shortstack{\textbf{Sent.}\\\textbf{Sim.}} & \textbf{B1}    & \textbf{B2}    & \textbf{B3}    & \textbf{B4}    & \textbf{METEOR} & \textbf{ROUGE}  \\
        \midrule
        \textit{ORG}  & 86.28 & 76.74 & 40.34 & 8.49 & 2.54 & 1.15 & 17.57 & 19.09\\
        \textit{AUG} & 85.72 & 71.91 & 32.54 & 4.02 & 0.69 & 0.17 & 13.84 & 11.07 \\
        \textit{FULL} & 85.74 & 72.42 & 32.95 & 4.19 & 0.85 & 0.25 & 14.16 & 11.79  \\
        \bottomrule
    \end{tabular}
    % }
    \caption{Automatic evaluation of follow-up question generation with human reference. \textit{ORG} (baseline) performs slightly better.}
    \label{tab:auto_eval_ref}
\end{table}  

\paragraph{Similarity to the references.} To compare our results with those obtained by \citet{meng-etal-2023-followupqg}, we perform identical automatic evaluations. We measure lexical overlap with BLEU-1-4 \citep{papineni-etal-2002-bleu}, METEOR \citep{lavie-agarwal-2007-meteor}, and ROUGE-L \citep{lin-2004-rouge}, and semantic similarity with BERTScore \citep{bert-score} and an embedding-based cosine score computed with \texttt{all-mpnet-base-v2} \citep{reimers-gurevych-2019-sentence}, following \citet{meng-etal-2023-followupqg}. 
For each metric, we compute the highest score across all generated follow-ups with the human reference and report the average for the entire dataset.  
Table~\ref{tab:auto_eval_ref} shows a consistent advantage for \textit{ORG}.  This is expected, as both its training data and the test set come from the original \fd{} distribution.  The lower BLEU scores for \textit{AUG} and \textit{FULL} reflect a common issue in open-ended QG: lexically diverse yet valid questions are under-rewarded by n-gram metrics \citep{pan-etal-2021-zero}.  In contrast, the gap between \textit{FULL} and \textit{AUG} is much smaller on BERTScore and embedding similarity, which focus on semantic alignment and discount stylistic differences. Consequently, we turn to human evaluation to capture diversity and nuanced informativeness that automatic metrics may overlook.

\begin{table}[b]
\centering
\scriptsize
\setlength{\tabcolsep}{3pt}
\begin{tabular}{p{0.2\columnwidth} p{0.35\columnwidth} p{0.35\columnwidth}} 
\toprule
% \addlinespace[0.1cm]
 & \textbf{Question} & \textbf{Numeric Scale} \\ 
 % \addlinespace[0.1cm]
 \midrule
% \addlinespace[0.2cm]
Validity & Is the FQ question a valid question? & yes (1) / no (0) \\
\cmidrule(lr){2-3}
 & Does the FQ contain any of the following errors? 
 & contains errors: \textbullet~redundant \textbullet~repetitive \textbullet~wrong semantic collocation (1) / no errors (0) \\ 
\midrule
Complexity & Does generating the FQ require reasoning? & complex (3) / moderate (2) / minimal (1) / no (0) \\ 
\midrule
Relevance & How relevant is FQ to the initial question and answer? & strongly (3) / relevant (2) / slightly (1) / not (0) \\ 
\midrule
Informativeness & Does the FQ elicit new information? & a lot (3) / some (2) / little (1) / no (0) \\ % for the audience? \\ 
\bottomrule
 \end{tabular}
\caption{The aspects evaluated in the human evaluation with respect to the follow-up question (FQ).}
\label{tab:annotation_questions}
\end{table}

\subsection{Human Evaluation}
\label{sec:results:human}
\begin{table}[!t] % Use table* to span both columns
    \centering
    \scriptsize
    \setlength{\tabcolsep}{2pt}  % Tighter column padding
    % \renewcommand{\arraystretch}{1.3}
    % \resizebox{\textwidth}{!}{ % Ensures table fits across both columns
        \begin{tabular}{ lcccccc }
            \toprule
             &  
            \multicolumn{2}{c}{\textit{ORG}} & 
            \multicolumn{2}{c}{\textit{AUG}} &
            \multicolumn{2}{c}{\textit{FULL}}\\
            \cmidrule(lr){2-3} \cmidrule(lr){4-5} \cmidrule(lr){6-7}
            & \textbf{Mean} & \textbf{Variance} & \textbf{Mean} & \textbf{Variance} & \textbf{Mean} & \textbf{Variance} \\
            \midrule
            Validity & 0.7324 & 0.1964 & \textbf{0.9065*} & 0.0849 & 0.8743 & 0.1102 \\
            % Error & 0.1769 & 0.1459 & 0.1963 & 0.1582 & \textbf{0.1393} & 0.1203 \\
            Complexity & 0.9274 & 1.0129 & \textbf{1.4798*} & 0.9441 & 1.4454 & 0.7025 \\
            Relevance & 1.6236 & 1.4716 & \textbf{2.0935*} & 1.0225 & 1.7377 & 1.0269\\
            Informativeness & 0.7755 & 0.9563 & \textbf{1.4517*} & 1.1297 & 1.2951 & 0.8223\\
            \bottomrule
        \end{tabular}
    % }
    % \captionsetup{justification=centering}
    \caption{Human evaluation scores for each aspect comparing \textit{ORG}, \textit{AUG} and \textit{FULL}. Best results are in bold; only statistically significant results are marked with an asterisk.}
    \label{tab:mean_variance}
\end{table}
\begin{table}[!b] % Use table* to span both columns
    \centering
    \scriptsize
    \setlength{\tabcolsep}{2pt}  % Tighter column padding
    % \renewcommand{\arraystretch}{1.3}
    % \resizebox{\textwidth}{!}{ % Ensures table fits across both columns
        \begin{tabular}{ lcccccc }
            \toprule
             &  
            \multicolumn{2}{c}{\textit{Teacher}} & 
            \multicolumn{2}{c}{\textit{AUG}} &
            \multicolumn{2}{c}{\textit{CoT}}\\
            \cmidrule(lr){2-3} \cmidrule(lr){4-5} \cmidrule(lr){6-7}
            & \textbf{Mean} & \textbf{Variance} & \textbf{Mean} & \textbf{Variance} & \textbf{Mean} & \textbf{Variance} \\
            \midrule
            Validity & 0.9457 & 0.0514 & 0.9065 & 0.0849 & \textbf{0.9647} & 0.0342 \\
            Complexity & \textbf{1.6227} & 0.6604 & 1.4798 & 0.9441 & 1.5725 & 0.7575 \\
            Relevance & 2.1240 & 0.7928 & 2.0935 & 1.0225 & \textbf{2.4863*} & 0.5500\\
            Informativeness & 1.6744 & 0.6502 & 1.4517 & 1.1297 & \textbf{1.6784} & 0.8095\\
            \bottomrule
        \end{tabular}
    % }
    % \captionsetup{justification=centering}
    \caption{Human evaluation scores for each aspect comparing (i) \textit{AUG} with \textit{Teacher} and (ii) \textit{AUG} with \textit{COT} GPT-4o. Best results are in bold; only statistically significant results are marked with an asterisk.}
    \label{tab:baseline_variance}
\end{table}

We conducted a human evaluation to assess the quality of generated FQs on four criteria: validity, complexity (the level of reasoning required), relevance, and informativeness (Details listed in Table~\ref{tab:annotation_questions}). We randomly sampled 30 (IQ, IA) pairs from the \fd{} testing set and evaluated the generated FQs from all five models. 

Human evaluation was conducted on Cloud Connect. To ensure high-quality annotations, we restricted participation to native English-speaking annotators with a minimum of 1,000 completed annotation tasks and an approval rating exceeding 90\%. Annotators interacted with a structured evaluation interface (see Appendix~\ref{app:interface}). Each task presented an initial question, its corresponding answer, and a generated FQ candidate. Annotators first assessed whether the FQ was valid. If deemed invalid, they proceeded directly to the next task. Otherwise, they answered four additional evaluation questions, as detailed in Table~\ref{tab:annotation_questions} (See Appendix~\ref{Model Evaluation - Human Annotation Guideline} for the complete annotation guidelines). 
Each task was annotated by 3 annotators, yielding substantial inter-annotator agreement, with an average Cohen's Kappa of $\kappa = 0.77$ \cite{cohen1960coefficient}.

\begin{table*}[ht]
    \centering
    \footnotesize
    \renewcommand{\arraystretch}{1.2}  % Ensure row heights are consistent
    \begin{tabular}{m{0.05\textwidth} m{0.9\textwidth}}
        \toprule
        \textbf{IQ} & How would you explain the concept of a heuristic?\\
        \midrule
        \textbf{IA} & It took me a long time to understand heuristics and \textcolor{blue}{\textit{I still don't know if I can ELI5 after a handful of bachelor and graduate degrees.}}  A heuristic is like a clue or a key to help you solve a problem or remember something. That's the best I got. \\
        \midrule
        \textbf{CA} & A heuristic is a practical approach or mental strategy used for problem-solving and decision-making, particularly when dealing with complex or incomplete information. It serves as a cognitive shortcut, allowing individuals to reach quick and satisfactory solutions without the need for exhaustive analysis or perfect precision. By focusing on the most relevant aspects of a problem and often relying on past experiences, common sense, or general rules of thumb, heuristics simplify the decision-making process and reduce cognitive load. While heuristic methods are not guaranteed to provide the best solution, they are typically sufficient for achieving an immediate goal and are useful in many everyday situations where speed and practicality are prioritized over exhaustive accuracy. \\
        \midrule
        \shortstack{\textbf{FQ}\\by\\\textit{ORG}}  & 
            -- \textcolor{red}{\textit{How do you know if you can ELI5?}}

            -- \textcolor{red}{\textit{What is a heuristic?}}
            
            -- Thanks for the answer.  So, what I am trying to understand is, \textcolor{red}{\textit{what is ELI5?}}
            
            -- \textcolor{red}{\textit{That's the best I got. I'm not a psychologist, but my brain picks out heuristics when I'm stressed and it usually works. I don't know if I can ELI5 after a handful of bachelor and graduate degrees. But I'm still working on it.}}
 \\
        \midrule
        \shortstack{\textbf{FQ}\\by\\\textit{AUG}} &
        -- What are some examples of cognitive strategies that rely on heuristics?
        
        -- How does the concept of a heuristic relate to the process of problem-solving?
        
        -- What role does the hippocampus play in heuristic formation?
\\
        \midrule
        \shortstack{\textbf{FQ}\\by\\\textit{FULL}} & 
        -- What are some examples of heuristics used in decision-making?
        
        -- What role do heuristics play in problem-solving?
        
        -- How do heuristics \textcolor{red}{\textit{help in segmenting and segmenting information}} to specific tasks or objectives?

        -- \textcolor{red}{\textit{I'm not sure I can ELI5 after a handful of bachelor and graduate degrees.  I'm not even sure if I can understand ELI4.   I know that it's a heuristic.  But I don't know if I understand ELII5.  Like, I know what a heuristics are.  And I know how to use a heymn to solve problems.  So I'm asking if you can ELII4?}}
\\
        \bottomrule
    \end{tabular}
    \caption{Example of follow-up question generated by three model variants, with comprehensive answers (ID 3182).}
    \label{tab:example_question_long}
\end{table*}

Table \ref{tab:mean_variance} reports the mean and variance of each evaluation criterion for \textit{ORG}, \textit{AUG}, and \textit{FULL}. Overall, \textit{AUG} achieved the best results across all criteria, with a statistically significant difference from the other models (tested with a one-way ANOVA). Over 90\% of the FQs generated by \textit{AUG} were considered valid, and these questions were rated as relevant, somewhat informative, and minimally to moderately complex. \textit{FULL} closely follows across aspects, while \textit{ORG} lags behind. The only aspect on which \textit{ORG} closely follows \textit{FULL} is relevance, aligning with the findings of \citet{meng-etal-2023-followupqg} that current models perform well in maintaining relevance. In sum, the results clearly prefer the questions generated by \textit{AUG}, which excel in validity, complexity, relevance, and informativeness---qualities essential for meaningful follow-up questions.

% COT/GPT model discussion
To gauge how effectively our information-gap pipeline distills LLM knowledge into smaller models, we repeated human evaluation on the GPT-4o \textit{Teacher} and \textit{CoT} baselines described in \S\ref{sec:exp_setup}. Table \ref{tab:baseline_variance} shows that both LLM baselines post slightly higher mean scores than the distilled \textit{AUG} model, yet the difference is statistically significant only for the \textsc{Relevance} metric. On validity, complexity, and informativeness, \textit{AUG}---a BART-large model trained on a random, size-matched subset of our augmented data---remains statistically indistinguishable from the much larger \textit{Teacher} and \textit{CoT} models. These results underscore the strength of our pipeline: by contrasting initial answers with LLM-generated comprehensive answers, it simulates the human information-seeking process and produces synthetic follow-up questions rich in diversity and informativeness. Fine-tuning on this augmented data enables small, locally deployable models to reach teacher-level quality at a fraction of the inference cost, thereby making high-performance \fd{} feasible on resource-constrained hardware.

The comparative results across models reveal key insights into the role of data quality versus quantity in the task of \fd. Notably, \textit{AUG}, trained on the same number of instances as \textit{ORG} but consisting solely of GPT-4o-generated, high-quality, reasoning-heavy questions, consistently outperforms both \textit{ORG} and \textit{FULL} across most metrics, yielding greater validity, complexity, relevance, and informativeness. This indicates that data quality is more critical than dataset size. Despite consisting of ten times more training data, \textit{FULL} failed to surpass \textit{AUG}, likely because remaining lower-quality or low-reasoning examples from the original dataset dilute learning. These findings challenge the assumption that larger datasets automatically improve performance and underscore the value of targeted augmentation with strict quality control. Future work should explore strategies to scale data while maintaining rigorous curation to further advance follow-up question generation.

\section{Analysis}
\label{sec:analysis}
To further understand the strengths and limitations of our method, we present a qualitative comparison of follow-up questions generated by all models for the same (IQ, IA) pair (\S\ref{sec:analysis:casestudy}), as well as an analysis of the expected information gain from the generated follow-up questions (\S\ref{sec:analysis:infogain}). 

\subsection{Qualitative Analysis}
\label{sec:analysis:casestudy}

In Table \ref{tab:example_question_long}, we compare follow-up questions generated by the \textit{ORG}, \textit{AUG}, and \textit{FULL} models for a given (IQ, IA) pair. Questions from \textit{ORG} are often redundant---e.g., ``What is a heuristic?''---or tangential, such as ``How do you know if you can ELI5?'' to the original responder that mentioned they didn't know if they could explain it to a 5-year-old (ELI5), thus drifting away from the target concept of heuristics. While the \textit{FULL} model yields a wider range of relevant questions and excels in diversity, it occasionally produces tangential or wordy phrasing, for instance, ``How do heuristics help in segmenting and segmenting information for specific tasks?'', which hurts clarity. By contrast, \textit{AUG} strikes the best balance of informativeness and diversity, offering focused, insightful questions like ``What are some examples of cognitive strategies that rely on heuristics?'' and ``How does the concept of a heuristic relate to the process of problem-solving?''. Additional examples can be found in Appendix \ref{app:additional_examples}.

\subsection{Quantifying Information Gain}
\label{sec:analysis:infogain}

In \S\ref{sec:results:human} we asked annotators to rate the informativeness of each follow-up question. We now introduce an automated alternative that requires no human raters, leveraging the GPT-4o ``comprehensive answers'' (CA; see definition in \S\ref{sec:data:augmentation}). We treat each CA as a proxy for the full body of information relevant to its (IQ, IA) context. An FQ is informative if it (i) cannot be answered from the IA alone---otherwise it adds no new information---and (ii) can be answered from the CA---otherwise it is likely irrelevant. Guided by this rule, we prompt GPT-4o to judge the answerability of every model-generated FQ against both the IA and the corresponding CA.

\begin{table}[!t]
    \centering
    \footnotesize
    \setlength{\tabcolsep}{3pt} % Adjust column spacing
    \resizebox{\linewidth}{!}{ 
    \begin{tabular}{lccc}
        \toprule
        \textbf{Model} & \textbf{Human-INF} & \textbf{GPT-INF-All (\%)} & 
        \textbf{GPT-INF-Sel (\%)}\\
        \midrule
        \textit{ORG}  & 0.7755 & 25.17 & 23.29\\
        \textit{AUG}  & \textbf{1.4517} & \textbf{36.19} & \textbf{35.91}\\
        \textit{FULL} & 1.2951 & 34.90 & 32.20 \\
        \bottomrule
    \end{tabular}
    }
    \caption{Comparison of human-annotated informativeness scores and GPT-evaluated informative percentage across models.}
    \label{tab:auto_informative}
\end{table}

% 0.7755 & 0.9563 & \textbf{1.4517*} & 1.1297 & 1.2951 & 0.8223\\
Table \ref{tab:auto_informative} corroborates the human evaluation of informativeness: \textit{AUG} produces the largest share of informative questions (36 \%), followed by \textit{FULL} (35 \%) and \textit{ORG} (25 \%). Comparing the GPT-4o labels with human-annotated informativeness scores (\S\ref{sec:results:auto}) further validates the automatic method: annotators assigned higher mean scores to FQs the model classified as informative (1.29) than to those it did not (1.07). A two-sample \textit{t}-test ($p = 0.0011$) confirms the statistical significance, although the effect size is small (Cohen's $d = 0.215$) \cite{cohen2013statistical}.

\section{Conclusion}
\label{sec:conclusion}
In this work, we proposed a novel approach to enhance the diversity and informativeness of follow-up questions by explicitly modeling information gaps via an LLM-generated comprehensive answer. We augmented the original \fd{} dataset with GPT-4o and distilled this data into a small, locally deployable BART-large model. Experiments show that our pipeline enables the small model to outperform peer small-model baselines and to perform comparably to GPT-4o baseline models in terms of validity, complexity, relevance, and informativeness---all at a fraction of the inference cost. These results suggest that targeted, high-quality augmentation can be more impactful than merely increasing dataset size. They also demonstrate that our method offers a practical approach for improving information-seeking dialogues---by reducing ambiguities and enhancing LLM responses---even on systems with limited computational resources.

Future work could explore ways to increase follow-up-question diversity while reducing redundancy, and to extend the pipeline to downstream tasks involving multi-turn dialogue. We also encourage research on stronger automated metrics for evaluating question quality, given the high cost of human annotation and the limitations of current automatic measures.

\section*{Limitations}
\label{sec:limitations}
We acknowledge several limitations in our work. First, while our CA-based pipeline is effective in knowledge-driven contexts, its applicability to non-knowledge-based conversations, such as opinion-based questions (e.g., ``What would you do in such a scenario?''), remains unclear, as the subjective judgment required in these conversations can be difficult for a generated CA to capture. Additionally, although our pipeline prioritizes informativeness, follow-up questions do not always need to introduce new information \cite{kurkul2018question}---for example, requests for simpler explanations (e.g., ``Can you explain this in an easier-to-understand way?''). 

Our work also calls for several future works and expansions. For example, our pipeline can be tested and evaluated on languages besides English, including low-resource languages. Moreover, given the pragmatic applicability of this pipeline and its focus on resource-efficiency, it would be pertinent to evaluate the compute-cost tradeoffs to help users make informed decisions. Lastly, our pipeline's performance can be evaluated on different combinations of Teacher-Student models as well. In the future, we hope to extend this method to support various types of follow-up questions and integrate it into downstream dialogue-based applications.

\section*{Ethical Considerations}
\label{sec:ethics}
All annotators involved in the human evaluation for this research were fairly compensated, with payment rates exceeding the local minimum wage to ensure equitable remuneration for their time and effort. Prior to recruiting annotators, ethical approval was obtained from the research ethics board at the authors’ institution, ensuring that the human evaluation process adhered to ethical guidelines and that no harm was caused to any individual involved. Additionally, the \fd{} dataset used in this work is publicly available, and we also released the new data created in this work, including the augmented data and generated comprehensive answers, to promote transparency and reproducibility in future work.

\section*{Acknowledgments}
This work was funded, in part, by the Vector Institute, Canada CIFAR AI Chairs program, Accelerate Foundation Models Research Program Award from
Microsoft, and an NSERC discovery grant. 

\newpage

% \bibliographystyle{acl_natbib}
\bibliography{anthology, custom}

\newpage
\appendix

\section{Problematic and Unrelated Instance Examples}
\label{app:problematic_sample}
Tables \ref{tab:task_sample_1696} and \ref{tab:task_sample_641} show examples of a problematic instance and an unrelated instance from \fd{} training set, respectively.
\begin{table}[H]
    \centering
    \scriptsize
    \renewcommand{\arraystretch}{1.3}
    \resizebox{\linewidth}{!}{ 
        \begin{tabular}{p{0.25\linewidth} | p{0.75\linewidth} }
            \toprule
            \textbf{Initial Question} & How exactly do you die if you touch electric wires? Do you burn? Suffocate? Heart stop? \\
            \midrule
            \textbf{Initial Answer} & If the current going thru you is around 1 amp, the shock may give you a heart attack. So it's like the hospital shock paddles, only it kills you. If the current is like 10 amps, it probably won't give you a heart attack, and you may survive if your organs aren't fully cooked. Which they definitely could be if the current isn't immediately stopped. And cooked kidneys don't function that good. So to summarize, either heart attack or cooked organs/burned, depending on the current. Edit: To clarify, it takes less than 1 amp to stop your heart, and the current also has to go thru your heart. I just used 1 amp and 10 amps as sample values. \\
            \bottomrule
        \end{tabular}
    }
    \caption{Problematic Instance 1696: Self-Harm-Related Content.}
    \label{tab:task_sample_1696}
\end{table}

\begin{table}[H]
    \centering
    \scriptsize
    \renewcommand{\arraystretch}{1.3}
    \resizebox{\linewidth}{!}{ 
        \begin{tabular}{ p{0.25\linewidth} | p{0.75\linewidth} }
        \toprule
        \textbf{Initial Question} & How its possible for the light to travel constantly with out stop slowing down from the most old galaxies? \\ \midrule
        \textbf{Initial Answer} & Okay, an NFT is a NON-FUNGIBLE TOKEN. Let's take the Mona Lisa for example. [...] So what you've done is bought the off-brand version (online) instead of the real thing. Also, the guy you paid doesn't really own the Mona Lisa. The TLDR is that you're buying an online version of a product and it says you own it, but you don't own the real thing.  \\ 
        \midrule
        \textbf{Follow-up Question} & Thank you for this!  My question is, what's the appeal?  Is it just rich-guy-bragging-rights? [...]  With, say, a baseball card you can display it and know that it's a rare item, and someday resell it at (you hope) a profit.  Are some of these people just speculating and hoping the `value' goes up? \\ 
        \bottomrule
        % You say to a guy, ``hey, I wanna buy the Mona Lisa!'' and this guy says ``sure, pay X amount and I'll hand you a receipt''. So, you pay X amount and you get this lil receipt that says you own the Mona Lisa. That's stored away in a cupboard in amongst all of these other cupboards, except the cupboards are the internet. However, you ask for the Mona Lisa because you've bought it, but this guy is like ``no, you don't own the REAL Mona Lisa, you own the Mona Lisa''.
    \end{tabular}
    }
    \caption{Unrelated Instance 641.}
    \label{tab:task_sample_641}
\end{table}

\section{LLM Prompts}
\label{app:prompts}

\begin{table}[H]
    \centering
    \scriptsize
    \renewcommand{\arraystretch}{1.3}
    \resizebox{\linewidth}{!}{ 
        \begin{tabular}{  p{0.25\linewidth} | p{0.75\linewidth} }
            \toprule
            \ & \textbf{Prompts} \\
            \midrule
            \textbf{Initial Answer} & ``Generate an answer focused on a single perspective only, without any conversational fillers. Do not repeat the question in the answer.'' \\
            \midrule
            \textbf{Next Answer} & ``Please provide a new answer focused on a different perspective, ensuring no overlap with previous answers. Focus on unique aspects or insights not covered earlier, and provide the answer only without any conversational fillers. Do not repeat the question in the answer.'' \\
            \midrule
            \textbf{Comprehensive Answer} & ``Synthesize the following answers into a single, comprehensive response. Integrate the key points and insights from each answer, ensuring a cohesive and well-rounded explanation. The final answer should be thorough and address multiple aspects of the question without unnecessary repetition.'' \\
            \bottomrule
        \end{tabular}
    }
    \caption{Comprehensive Answer Generation Prompts: GPT-4o first generates an answer from a single perspective, then iteratively provides non-overlapping answers from different perspectives, which are finally synthesized into a unified response.}
    % LLM was sequentially prompted to generate answers with uncovered information, then compiled into a comprehensive answer.
    \label{tab:comprehensive_answer_generation}
\end{table}

\begin{table}[H]
    \centering
    \footnotesize
    \renewcommand{\arraystretch}{1.3}
    \resizebox{\linewidth}{!}{ % Adjust table width to fit the document
        \begin{tabular}{ p{\columnwidth} }
            \toprule
            \textbf{Information Gap Identification \& Follow-up Question Generation} \\
            \midrule
        ``Generate all possible follow-up questions as candidates. These follow-up questions must be related to the original question, but must not be rephrases of the original question. These follow-up questions should be answerable by the complete answer. These follow-up questions should not be answered, covered, or detailed by the original answer, but must target terminologies mentioned in the original answer. Separate each follow-up question with `<sep>`.'' \\
            \bottomrule
        \end{tabular}
    }
    \caption{Follow-up Question Generation Prompt.}
    \label{tab:followup_generation_prompt}
\end{table}

\section{Augmented Data - Human Annotation Guideline}
\label{sec:augmented_data_annotation_guideline}

Table \ref{tab:task_description_GPT} presents the job description and annotation questions for our human annotation task.

\begin{table}[H]
    \scriptsize
    \centering
    \begin{tabular}{ p{\columnwidth} }
        \toprule
        \textbf{Job Description} \\
        \midrule
        Welcome, and thank you for participating in this text evaluation task! In this job, you'll be helping us verify the quality of follow-up questions generated by GPT.\\
        For each task, we will provide you with a pair consisting of a question and answer collected from Reddit's ``Explain Like I'm Five'' (ELI5) forum. You will be asked to evaluate the quality of the follow-up question generated by GPT. These questions and answers aim to provide layperson-friendly explanations for real-life queries. Here is an example of one task sample: \\
        Each task may contain noise, such as invalid follow-up questions, sensitive information, or questions unrelated to the original question or answer. Your role is to help us identify these noisy samples.\\
        For each task, you will be shown one triple (question, answer, follow-up question). Carefully review each component and answer the following questions based on your judgment: \\
        \midrule
        \textbf{Q1:} Do you think the follow-up question is a valid question? \\
        \textbf{A.} Yes \quad
        \textbf{B.} No \\
        \midrule
        \textbf{Q2:} Does the initial question, answer, or follow-up question contain sensitive information? \\
        \textbf{A.} Yes \quad
        \textbf{B.} No \\
        \midrule
        \textbf{Q3:} Do you think the follow-up question is related to the original question and the answer? \\
        \textbf{A.} Strongly Related \quad
        \textbf{B.} Related \quad
        \textbf{C.} Slightly Related \quad
        \textbf{D.} Not Related \\
        \bottomrule
    \end{tabular}
    \caption{Task description and evaluation questions used for human annotation of augmented data.}
    \label{tab:task_description_GPT}
\end{table}

\subsection{Valid/Invalid Question Guideline}

The follow-up question might contain multiple sentences but it should consist of at least one valid question. A valid question must be in a question format and ask meaningful information, including Wh-questions (what/why/where/etc.), open-ended questions, probing questions and etc. Invalid questions like ``10000 meters? really?'', are often used in conversational speech to express feelings instead of asking for new information. 
Table~\ref{tab:valid_invalid_followupq} contains examples of valid and invalid follow-up questions.

\begin{table}[H]
    \centering
    \small
    \renewcommand{\arraystretch}{1.3}
    \resizebox{\linewidth}{!}{ 
        \begin{tabular}{ p{0.5\linewidth} | p{0.5\linewidth} }
            \toprule
            \multicolumn{2}{p{\linewidth}}{\textbf{Initial Question:} Why is the sea calm in the mornings?} \\
            \midrule
            \multicolumn{2}{p{1.05\linewidth}}{\textbf{Initial Answer:} There are two types of waves which can turn a flat sea into a rougher one - swell waves and wind waves. Swell waves can arrive at any time of day, but because wind waves are generated by the wind, they only develop when the wind begins to blow steadily. Since wind speeds are often low at night, and increase during the daytime, wind waves often die out during the night, leading to a relatively flat sea (perhaps with swell waves) in the early morning. During the day, the wind waves increase in size as the wind speed increases, leading to a rougher, more choppy, sea surface during the afternoon and evening. } \\
            \midrule
            \textbf{Valid Follow-up} & \textbf{Invalid Follow-up} \\
            \midrule
            Why are winds always weak in the morning and very strong during the day? & Isn't it common sense that the sea is calmer in the morning?\\
            \midrule
            \textbf{Reason} & \textbf{Reason}\\
            \midrule
            The follow-up question is a ``Why'' question, asking specific reasons about the change of the winds. Therefore, it is a valid question.  & This is a rhetorical question because it does not genuinely seek new information. It implies that the answer is obvious and does not contribute to the discussion. \\
            \bottomrule
        \end{tabular}
    }
    \caption{Examples of valid and invalid follow-up questions. For the given initial question and answer, the left column presents a valid follow-up question, while the right column features an invalid one, each accompanied by corresponding reasons below.}
    \label{tab:valid_invalid_followupq}
\end{table}

\subsection{Inappropriate Question Guideline}

Examples of racist comments include: ``It's credit to your race,'' ``Black people will not understand.'' Examples of hate speech include: ``He should go back to where he comes from,'' ``All Mexicans are rapists.'' Examples of offensive or rude comments include: ``Women are not suitable for working in the IT field,'' ``Gay will never understand.'' Table~\ref{tab:inappropriate_followupq} contains an example of an inappropriate follow-up question.

\begin{table}[H]
    \centering
    \small
    \renewcommand{\arraystretch}{1.3}
    \resizebox{\linewidth}{!}{ 
        \begin{tabular}{ p{0.5\linewidth} | p{0.5\linewidth} }
            \toprule
            \multicolumn{2}{p{\linewidth}}{\textbf{Initial Question:} Why do people develop eating disorders?} \\
            \midrule
            \multicolumn{2}{p{1.05\linewidth}}{\textbf{Initial Answer:} Eating disorders are complex mental health conditions influenced by a combination of genetic, psychological, environmental, and social factors. While societal beauty standards and pressures can contribute, eating disorders are not simply about wanting to be thin. Conditions like anorexia, bulimia, and binge-eating disorder involve intricate relationships between self-image, emotional regulation, and biological predispositions. Many individuals with eating disorders struggle with anxiety, depression, or trauma, which can further complicate their relationship with food.} \\
            \midrule
            \textbf{Inappropriate Follow-up} & \textbf{Reason} \\
            \midrule
            Why don't people with eating disorders just stop starving themselves and eat normally like everyone else? & This question is dismissive. The phrasing is insensitive and could be harmful to individuals struggling with these conditions.\\
            % \midrule
            % \textbf{Reason} & \textbf{Reason}\\
            % \midrule
            % The follow-up question is a ``Why'' question, asking specific reasons about the change of the winds during the day. Therefore, it is a valid question.  & This is a rhetorical question because it does not genuinely seek new information. Instead, it implies that the answer is already obvious and does not contribute to the discussion in a meaningful way. \\
            \bottomrule
        \end{tabular}
    }
    \caption{Example of an inappropriate follow-up question for the given initial question and answer, accompanied by corresponding reasons below.}
    \label{tab:inappropriate_followupq}
\end{table}

\subsection{Relevance Question Guideline}

\begin{itemize}
    \item \textbf{Strongly Related}: The follow-up question asks for specific definitions, particular reasons, or meanings directly from the original question and answer.
    \item \textbf{Related}: The follow-up question primarily seeks information from the original question or answer but also brings in additional, new information.
    \item \textbf{Slightly Related}: The follow-up question mainly addresses other cases but has some relevance to the original question or answer.
    \item \textbf{Not Related}: The follow-up question does not relate to the original question or answer.
\end{itemize}
Table~\ref{tab:relate_followupq} contains follow-up questions with various levels of relevance.

\begin{table}[H]
    \centering
    \small
    \renewcommand{\arraystretch}{1.3}
    \resizebox{\linewidth}{!}{ 
        \begin{tabular}{ p{0.5\linewidth} | p{0.5\linewidth} }
            \toprule
            \multicolumn{2}{p{\linewidth}}{\textbf{Initial Question:} Why do airplanes leave white trails in the sky?} \\
            \midrule
            \multicolumn{2}{p{1.05\linewidth}}{\textbf{Initial Answer:} Those white trails are called contrails, short for condensation trails. They form when hot exhaust from the airplane's engines mixes with the cold air in the upper atmosphere. The water vapor in the exhaust condenses and freezes into tiny ice crystals, creating the white streaks you see in the sky. The persistence of these trails depends on humidity levels; if the air is dry, the contrail dissipates quickly, but if the air is humid, the contrail can linger for a long time.} \\
            \midrule
            \textbf{Strongly Related Question Example} & \textbf{Related Follow-up Question Example} \\
            \midrule
            Why do some contrails last longer than others? & Do contrails have any impact on the environment?\\
            \midrule
            \textbf{Reason} & \textbf{Reason}\\
            \midrule
            The follow-up question directly builds on the information provided in the answer, specifically regarding the persistence of contrails. Since the answer already mentions humidity as a factor, this question seeks further clarification, making it strongly related.  & This follow-up question extends the topic of contrails by asking about their environmental impact. While the original answer does not discuss environmental effects, the question is still relevant because it builds on the phenomenon explained. Thus, it is considered related. \\
            \midrule
            \textbf{Slightly Related Question Example} & \textbf{Not Related Follow-up Question Example} \\
            \midrule
            Why do some airplanes make more noise than others? & What causes volcanoes to erupt?\\
            \midrule
            \textbf{Reason} & \textbf{Reason}\\
            \midrule
           The follow-up question is about airplanes, which is the general topic of the original question, but it shifts the focus from contrails to noise. While both topics are related to aviation, the connection between them is weak, making the question only slightly related.  & The follow-up question introduces a completely unrelated topic (volcanoes) that has no connection to airplanes, contrails, or atmospheric conditions. Since it does not build on the original question or answer in any way, it is considered not related. \\
            \bottomrule
        \end{tabular}
    }
    \caption{Examples of follow-up questions' relevance for the given initial question and answer, accompanied by corresponding reasons below.}
    \label{tab:relate_followupq}
\end{table}

\section{Baseline Reproduce}
\label{appendix:baseline_reproduce}
\begin{table*}[ht]
    \centering
    \footnotesize{
    \begin{tabular}{@{}lcccccccc@{}}
        \toprule
        & B1    & B2    & B3    & B4    & METEOR & ROUGE & BERT & Sentence Similarity \\ 
        \midrule
        \textit{Reported ORG} & 17.22 & 7.11  & 3.89  & 2.61  & 8.00   & 13.35 & - & -        \\ 
        \textit{Reproduced ORG}   & 21.88 & 3.27  & 1.62  & 1.13  & 14.76  & 12.39 & 84.41 & 70.73 \\ 
        \bottomrule
    \end{tabular}
    }
    \caption{Comparison of Results from Paper Report and Our Reproduction.}
    \label{tab:comparison}
\end{table*}
To establish a baseline, we attempted to reproduce the results of \citet{meng-etal-2023-followupqg} using the reported parameters, as the original implementation was unavailable. We use BART-large, consisting of 24 layers, 16 attention heads, and a hidden dimension of 1024. The initial learning rate (5e-5) led to training instability, which we mitigated by adjusting it to 2e-5 while keeping other hyperparameters unchanged (batch size: 8, epochs: 10, optimizer: Adam \cite{kinga2015method}). The training was conducted on an NVIDIA Tesla V100 GPU with 32GB of memory, taking approximately 6 hours per run. We set the random seed to 42. After multiple runs, our reproduced model yielded similar overall performance but with some variation-certain metrics improved while others slightly declined (see Table \ref{tab:comparison}). This model served as the baseline for subsequent experiments.

\section{Filtering Ungrammatical Follow-Up Questions: Examples and Pseudocode}
\label{appendix:ungrammatical_questions}

\begin{lstlisting}[language=Python, breaklines, basicstyle=\small\ttfamily]
def is_valid_question(FQ, IQ, IA):
    return (
        # confirm that FQ ends with '?'
        contains_question_mark(FQ) and 
        # use dependency parsing to check for
        # WH-Questions, Yes/No Questions
        # and Rhetorical Questions, 
        is_question_dependency_parsing(FQ) and 
        # checks for meaningless/invalid words
        # e.g. <QUS>, <EQT>
        not contains_invalid_word(FQ) and
        # checks for n-gram substring that is 
        # duplicated in both FQ and (IQ/IA)
        not contains_duplicate_words(IQ, IA, FQ)
    )
\end{lstlisting}

% \begin{minted}[breaklines, fontsize=\small]{python}
% def is_valid_question(FQ, IQ, IA):
%     return (
%         # confirm that FQ ends with '?'
%         contains_question_mark(FQ) and 
%         # use dependency parsing to check for
%         # WH-Questions, Yes/No Questions
%         # and Rhetorical Questions, 
%         is_question_dependency_parsing(FQ) and 
%         # checks for meaningless/invalid words
%         # e.g. <QUS>, <EQT>
%         not contains_invalid_word(FQ) and
%         # checks for n-gram substring that is 
%         # duplicated in both FQ and (IQ/IA)
%         not contains_duplicate_words(IQ, IA, FQ)
%         )
% \end{minted}

\begin{table}[H]
\centering
    \scriptsize
    \renewcommand{\arraystretch}{1.3}
    \resizebox{\linewidth}{!}{ 
        \begin{tabular}{  p{0.25\linewidth} | p{0.75\linewidth} }
        \toprule
        \textbf{Error Type} & \textbf{Example} \\
        \midrule
        Improper delimiter & \textit{> Constant reflux can cause cancer of the sphincter.<QUS> Is this true for people with GERD?} \\
        \midrule
        Non-question statement & \textit{I think that's the key. I'm not sure what you mean by ``3 components''. The rotation is defined by the third component. The fourth component is used to determine the degree of rotation. So, if you have a 3d vector, you can rotate it in any direction. But if you want to rotate it out to a 4d vector you need to use a fourth component.} \\
        \midrule
        Excessive context duplication & \textit{Now try doing it for tomorrow's price chart. When you look at the historical chart of a company's price you'll be able to pick good buying and selling points 100\% of the time. Now try to do it for today's price} Is that impossible? \\
        \bottomrule
    \end{tabular}
    }
    \caption{Follow-up Question Error Types and Examples.}
    \label{tab:error_types}
\end{table}

% \section{Additional Tables}
% \label{app:additional_tables}

% \subsection{Numerical Mapping of Human Evaluation Annotations}
% \label{app:numerical_mapping_of_human_eval}
% To enable quantitative analysis, we applied predefined mappings, as shown in Table~\ref{tab:user_selection_mapping}, to transform categorical annotations from human evaluation tasks into numerical representations. Binary values (0/1) were assigned to categorical judgments, such as the validity of follow-up questions (valid = 1, not valid = 0), while ordinal values (0-3) were used for graded attributes, such as relevance (not related = 0, strongly related = 3).

\section{Model Evaluation - Human Annotation Guideline}
\label{Model Evaluation - Human Annotation Guideline}

Table \ref{tab:task_description_bart} presents the job description and annotation questions for our human annotation task.

\begin{table}[H]
    \scriptsize
    \centering
    \begin{tabular}{ p{\columnwidth} }
        \toprule
        \textbf{Job Description} \\
        \midrule
        In this job, you'll be helping us evaluate the quality of follow-up questions generated by a language model called BART. \\
        For each task, we will provide you with a pair consisting of a question and answer collected from Reddit's ``Explain Like I'm Five'' (ELI5) forum. You will be asked to evaluate the quality of the follow-up question generated by BART. These questions and answers aim to provide layperson-friendly explanations for real-life queries. \\
        Our data may contain noise, such as invalid follow-up questions, errors, lack of reasoning, or follow-up questions unrelated to the original question or answer. Your role is to help us identify these noisy samples. \\
        For each task, you will be shown one triple (question, answer, follow-up question). Carefully review each component and answer the following questions based on your judgment: \\
        \midrule
        \textbf{Q1:} Do you think the follow-up question is a valid question? \\
        \textbf{A.} Yes \quad
        \textbf{B.} No \\
        \midrule
        \textbf{Q2:} How relevant is the follow-up question to the original question and answer? \\
        \textbf{A.} Strongly Related \quad
        \textbf{B.} Related \quad
        \textbf{C.} Slightly Related \quad
        \textbf{D.} Not Related \\
        \midrule
        \textbf{Q3:} Does the follow-up question contain any of the following errors? \\
        \textbf{A.} No Errors \quad
        \textbf{B.} Redundant \quad
        \textbf{C.} Repetitive  \quad
        \textbf{D.} Wrong Semantic Collocation \quad
        \textbf{E.} Other Errors \\
        \midrule
        \textbf{Q4:} Does generating this follow-up question require reasoning? \\
        \textbf{A.} Requires complex amount of reasoning \quad
        \textbf{B.} Requires moderate amount of reasoning \quad
        \textbf{C.} Requires minimal amount of reasoning \quad
        \textbf{D.} Does not require any reasoning \\
        \midrule
        \textbf{Q5:} Does the follow-up question contain new information for the audience? \\
        \textbf{A.} Introduces a lot of new information \quad
        \textbf{B.} Introduces some new information \quad
        \textbf{C.} Introduces little new information \quad
        \textbf{D.} Does not introduce any new information \\
        \bottomrule
    \end{tabular}
    \caption{Task description and evaluation questions used for BART follow-up question evaluation.}
    \label{tab:task_description_bart}
\end{table}

\subsection{Error Question Guideline}

\noindent Does the follow-up question contain any of the following errors?

\noindent \textbf{Identify any language issues in the follow-up question.}
\begin{itemize}
    \item \textbf{No Errors} – The follow-up question is appropriate and adds value.
    \item \textbf{Redundant} – The follow-up does not introduce any new information.
    \item \textbf{Repetitive} – The follow-up question closely mirrors the original question.
    \item \textbf{Wrong Semantic Collocation} – The question contains unnatural or incorrect phrasing.
    \item \textbf{Other Errors} – Any issues that do not fit the categories above.
\end{itemize}
Table~\ref{tab:error_followupq} contains examples of follow-up questions with various error status.

\begin{table}[H]
    \centering
    \small
    \renewcommand{\arraystretch}{1.3}
    \resizebox{\linewidth}{!}{ 
        \begin{tabular}{ p{0.5\linewidth} | p{0.5\linewidth} }
            \toprule
            \multicolumn{2}{p{\linewidth}}{\textbf{Initial Question:} How do vaccines work?} \\
            \midrule
            \multicolumn{2}{p{1.05\linewidth}}{\textbf{Initial Answer:} Vaccines work by training your immune system to recognize and fight specific germs. They contain harmless parts of the germ (or something similar) so that your body can learn to defend against it. This way, if you ever encounter the actual germ, your immune system can respond quickly and prevent illness.} \\
            \midrule
            \textbf{No Errors Example} & \textbf{Redundant Example} \\
            \midrule
            How does a vaccine train the immune system? & Are vaccines used to help the immune system recognize germs?\\
            \midrule
            \textbf{Reason} & \textbf{Reason}\\
            \midrule
           The follow-up question is well-formed, relevant, and adds value by diving deeper into a key concept from the original answer. It does not repeat information unnecessarily or contain any language errors.  & The follow-up question is redundant because it merely restates information already provided in the initial answer without adding depth or prompting new discussion. \\
            \midrule
            \textbf{Repetitive Example} & \textbf{Wrong Semantic Collocation Example} \\
            \midrule
            What do vaccines do? & Do vaccines memorize diseases?\\
            \midrule
            \textbf{Reason} & \textbf{Reason}\\
            \midrule
           This follow-up question is nearly identical to the original question, simply reworded. Since it does not introduce new angles or expand on any details, it is considered repetitive.  & The phrase``vaccines memorize diseases'' is unnatural and incorrect in this context. A better way to phrase the question would be: ``Do vaccines help the immune system remember diseases?'' \\
            \bottomrule
        \end{tabular}
    }
    \caption{Examples of follow-up questions' error status for the given initial question and answer, accompanied by corresponding reasonings below.}
    \label{tab:error_followupq}
\end{table}

\subsection{Reasoning Question Guideline}

Evaluate the level of reasoning needed to generate the follow-up question. 
\begin{itemize}
    \item \textbf{Complex reasoning} involves synthesizing multiple ideas or deeply analyzing information.
    \item \textbf{Moderate reasoning} requires interpreting the given content or slightly extending the discussion.
    \item \textbf{Minimal reasoning} involves simple comprehension or directly rephrasing information.
    \item \textbf{No reasoning} applies to questions that are direct repetitions or restatements without any thought process.
\end{itemize}
Table~\ref{tab:reason_followupq} contains examples of follow-up questions with various reasoning complexity.

\begin{table}[H]
    \centering
    \small
    \renewcommand{\arraystretch}{1.3}
    \resizebox{\linewidth}{!}{ 
        \begin{tabular}{ p{0.5\linewidth} | p{0.5\linewidth} }
            \toprule
            \multicolumn{2}{p{\linewidth}}{\textbf{Initial Question:} How does sleep affect brain function?} \\
            \midrule
            \multicolumn{2}{p{1.05\linewidth}}{\textbf{Initial Answer:} Sleep is essential for brain function because it helps with memory consolidation, cognitive processing, and emotional regulation. During sleep, the brain strengthens neural connections, removes toxins, and allows different areas to reset for the next day.} \\
            \midrule
            \textbf{Complex Amount of Reasoning Example} & \textbf{Moderate Amount of Reasoning Example} \\
            \midrule
            What are the long-term cognitive effects of chronic sleep deprivation compared to occasional sleep loss? & How does sleep remove toxins from the brain? \\
            \midrule
            \textbf{Reason} & \textbf{Reason}\\
            \midrule
           This follow-up question requires complex reasoning because it involves comparing two different scenarios (chronic vs. occasional sleep deprivation) and analyzing their distinct long-term effects on cognition, requiring deeper thought and synthesis of information.  & This follow-up question requires moderate reasoning because it builds on a specific detail from the original answer (toxin removal) and asks for an explanation of the biological process involved. \\
            \midrule
            \textbf{Minimal Amount of Reasoning Example} & \textbf{Does Not Require Any Reasoning Example} \\
            \midrule
            What are the benefits of sleep for memory? &  Does sleep help with memory?\\
            \midrule
            \textbf{Reason} & \textbf{Reason}\\
            \midrule
           This follow-up question requires minimal reasoning as it only asks for elaboration on a topic already stated in the original answer (memory consolidation), without introducing any new angle.  & This follow-up question does not require any reasoning since it directly repeats a fact already stated in the original answer, making it redundant. \\
            \bottomrule
        \end{tabular}
    }
    \caption{Examples of follow-up questions' reasoning complexity for the given initial question and answer, accompanied by corresponding reasons below.}
    \label{tab:reason_followupq}
\end{table}

\subsection{Informativeness Question Guideline}

Evaluate whether the follow-up question enriches the topic by providing or eliciting new information.

\begin{itemize}
    \item \textbf{A Lot of New Information} indicates a significant amount of new knowledge is introduced.
    \item \textbf{Some New Information} suggests moderate enrichment.
    \item \textbf{Little New Information} implies minimal addition.
    \item \textbf{No New Information} means no new information is provided to the audience.
\end{itemize}
Table~\ref{tab:inform_followupq} contains examples of follow-up questions with various informativeness levels.

\begin{table}[H]
    \centering
    \small
    \renewcommand{\arraystretch}{1.3}
    \resizebox{\linewidth}{!}{ 
        \begin{tabular}{ p{0.5\linewidth} | p{0.5\linewidth} }
            \toprule
            \multicolumn{2}{p{\linewidth}}{\textbf{Initial Question:} How do vaccines work?} \\
            \midrule
            \multicolumn{2}{p{1.05\linewidth}}{\textbf{Initial Answer:} Vaccines train the immune system to recognize and fight specific germs by introducing harmless parts of the germ or something similar. This prepares the body to respond quickly if exposed to the actual germ in the future.} \\
            \midrule
            \textbf{A Lot of New Information Example} & \textbf{Some New Information Example} \\
            \midrule
            What are the differences between traditional vaccines and mRNA vaccines? & How long does it take for a vaccine to provide immunity?\\
            \midrule
            \textbf{Reason} & \textbf{Reason}\\
            \midrule
           This follow-up question introduces a significantly new dimension by asking about different types of vaccines, which were not mentioned in the original answer, expanding the discussion substantially.  & The follow-up question adds moderately new information by focusing on the timeline of immunity development, a relevant but additional detail not covered in the initial answer. \\
            \midrule
            \textbf{Little New Information Example} & \textbf{Does Not Introduce Any New Information Example} \\
            \midrule
            Do vaccines help prevent disease outbreaks? & Do vaccines help the immune system recognize germs? \\
            \midrule
            \textbf{Reason} & \textbf{Reason}\\
            \midrule
           The follow-up question slightly expands the discussion by addressing disease outbreaks, but it is already implied in the original answer, as vaccines train the immune system to fight germs. & This follow-up question does not add any new information as it directly restates a key point from the original answer in slightly different words. \\
            \bottomrule
        \end{tabular}
    }
    \caption{Examples of follow-up questions' informativeness for the given initial question and answer, accompanied by corresponding reasons below.}
    \label{tab:inform_followupq}
\end{table}

\section{Additional Examples}
\label{app:additional_examples}
\begin{table*}[ht]
    \centering
    \small
    \renewcommand{\arraystretch}{1.2}  % Ensure row heights are consistent
    \begin{tabular}{m{0.05\textwidth} m{0.9\textwidth}}
        \toprule
        \textbf{IQ} & What causes occupational domination? \\
        \midrule
        \textbf{IA} & \textcolor{blue}{\textit{In IT it is cultural. Women in the U.S. drop out of IT classes because they do not like the male-dominated culture.}} Whereas India, Vietnam, and other countries where we outsource our IT have a much higher percentage of female programmers. \textcolor{blue}{\textit{It is not something about women.}} It is something about western IT classes and tech culture. \\
        \midrule
        \textbf{CA} & Occupation domination occurs due to a combination of historical, social, economic, and geographic factors. \textit{Historically}, certain groups have had greater access to education, resources, and opportunities, leading them to be more prevalent in specific jobs or industries. This is often reinforced by network effects, where individuals within a profession prefer to work with or hire people who share similar backgrounds, cultural norms, or experiences, thus perpetuating the concentration of certain demographics in those fields. \textit{Legal and institutional} barriers can also play a significant role. Discriminatory policies or practices can systematically exclude certain groups from specific professions, allowing other groups, who do not face such barriers, to dominate those fields. \textit{Geographic and environmental} factors further influence occupation domination. The presence of certain natural resources or particular climate conditions in an area can dictate the types of industries that develop there. Local populations often specialize in these industries, passing down specialized skills and knowledge through generations, which solidifies their dominance in those fields over time. \\
        \midrule
        \shortstack{\textbf{FQ}\\by\\\textit{ORG}}  &  
            -- But why do women drop out of IT classes?

            -- \textcolor{red}{\textit{In IT it is cultural. Women in the U.S. drop out of IT classes because they do not like the male-dominated culture.}} This is a myth. \textcolor{red}{\textit{It is not something about women.}} It's a myth that women drop out because of the male-dominant culture.
            
            -- \textcolor{red}{\textit{So it’s cultural?}}
            
            -- What about women who want to be in tech but are not in the workforce?
            
            -- Is there a cultural component of IT that makes it more difficult for women to break into it?
 \\
        \midrule
        \shortstack{\textbf{FQ}\\by\\\textit{AUG}} & 
        -- What are some examples of how outsourcing has impacted the gender imbalance in IT?
        
        -- How do cultural expectations influence the recruitment of female programmers?
        
        -- Can you explain the concept of `cultural fit' in the context of occupational domination?
\\
        \midrule
        \shortstack{\textbf{FQ}\\by\\\textit{FULL}} & 
        -- How do traditional gender roles in professions like medicine and engineering contribute to occupation domination?
        
        -- Why might women feel more comfortable pursuing professions outside of traditional professions?
        
        -- Can you explain the concept of `perceived value' in the context of occupation domination?
\\
        \bottomrule
    \end{tabular}
    \caption{Example of follow-up question generated by three model variants, with comprehensive answers (ID 3168).}
    \label{tab:example_question_short}
\end{table*}
See Tables~\ref{tab:example_question_short}

\section{Interface Examples}
\label{app:interface}
\begin{figure*}
    \centering
    \includegraphics[width=0.8\textwidth, keepaspectratio]{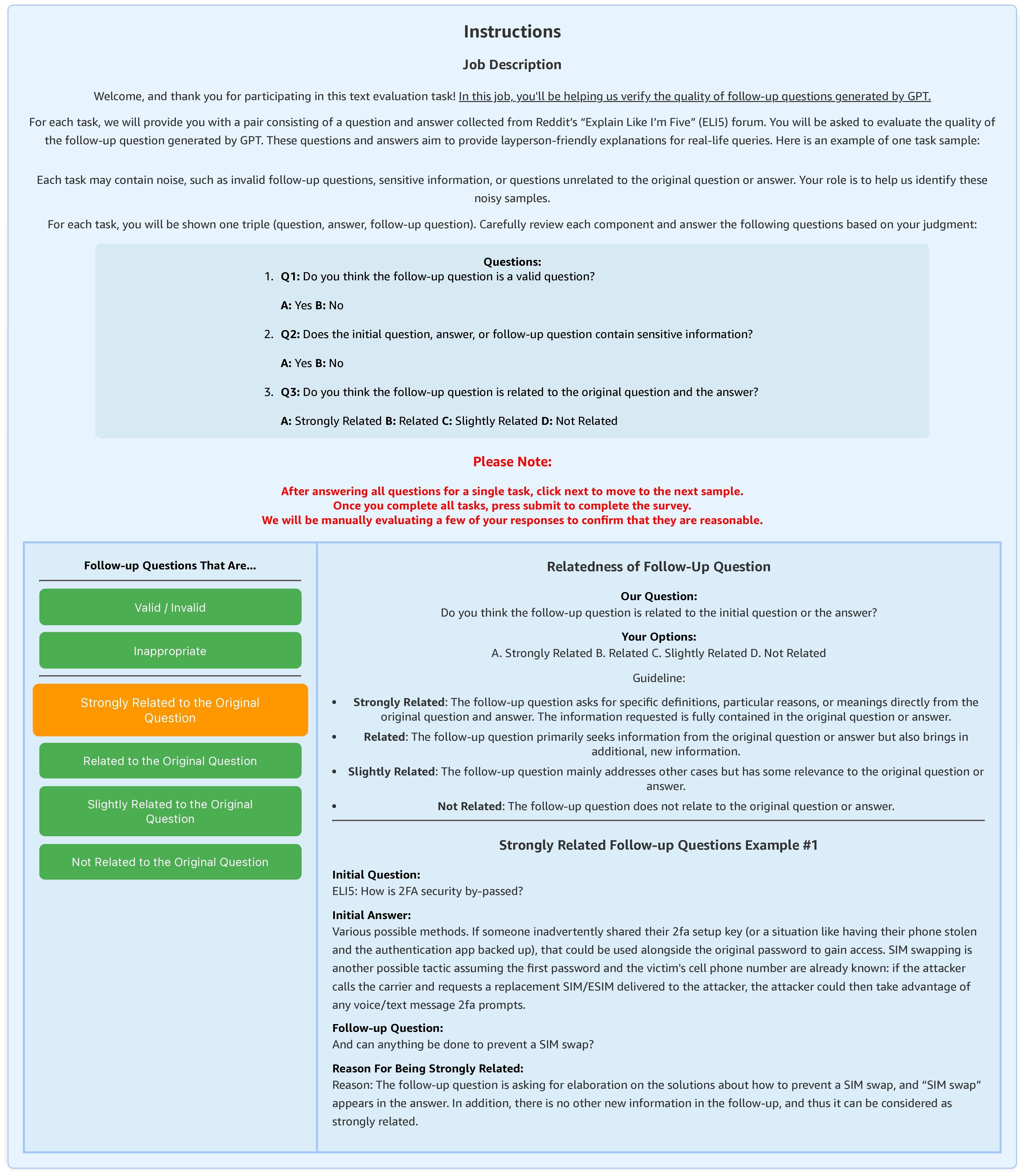}
    \caption{Human Evaluation Interface - Task Instructions and Examples.}
    \label{fig:instruction_interface}
\end{figure*}

\begin{figure*}
    \centering
    \includegraphics[width=0.8\textwidth, keepaspectratio]{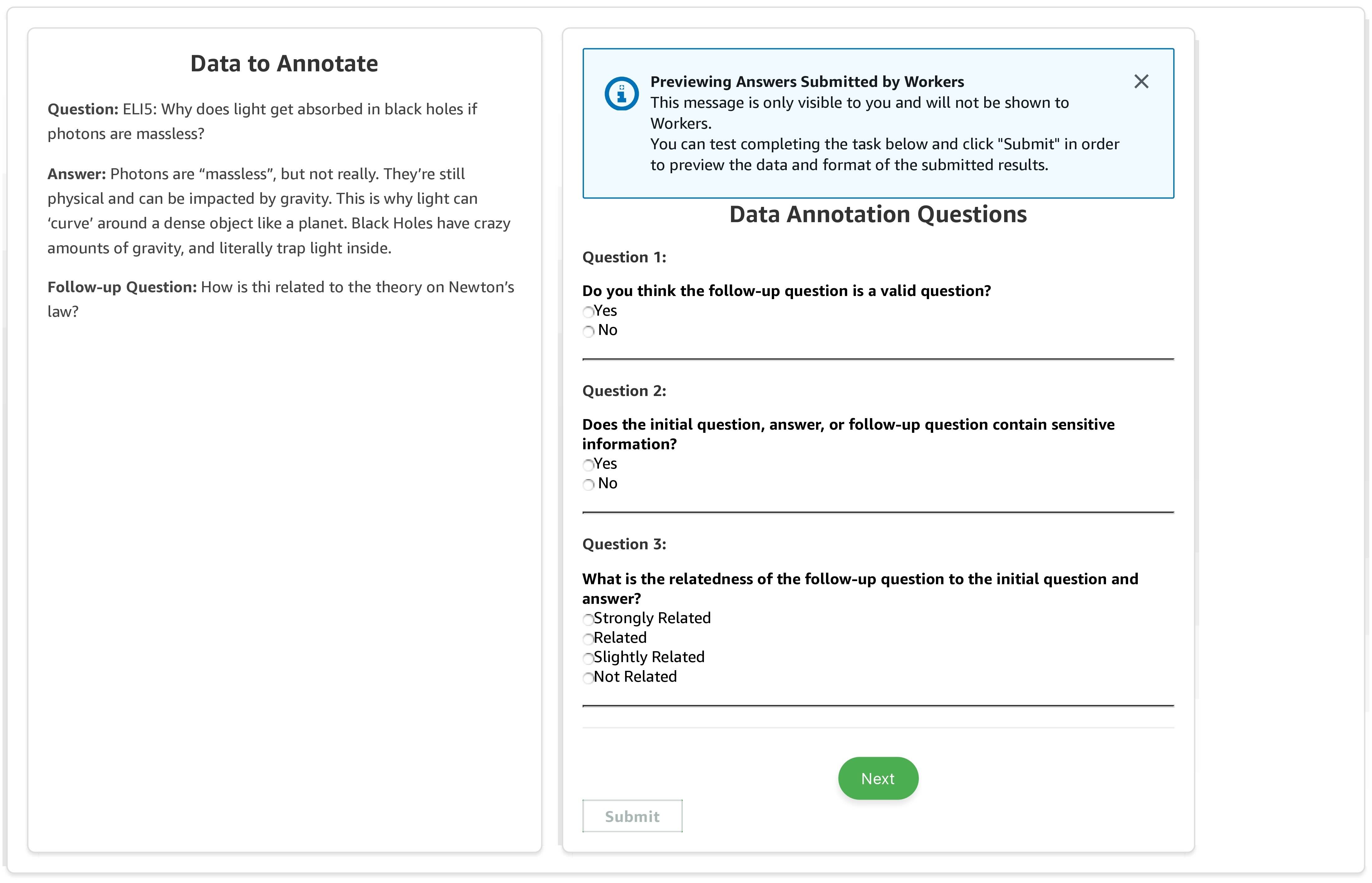}
    \caption{Human Evaluation Interface - Annotation.}
    \label{fig:annotation_interface}
\end{figure*}

See Figures~\ref{fig:instruction_interface} and \ref{fig:annotation_interface}

\end{document}